\documentclass[]{IEEEphot}
\usepackage{graphicx}
\usepackage{subcaption}
\usepackage{float}

\jvol{1}
\jnum{1}
\jmonth{Jan}
\pubyear{2025}

\begin{document}

\title{Inscanner:\\Dual-Phased Detection and Classification of\\ Auxiliary Insulation Using YOLOv8 Models}

\author{
    \IEEEauthorblockN{
        Youngtae Kim\IEEEauthorrefmark{1}, 
        Soonju Jeong\IEEEauthorrefmark{1}, 
        Sardar Arslan\IEEEauthorrefmark{1}, 
        Dhananjay Agnihotri\IEEEauthorrefmark{2}, 
        Yahya Ahmed\IEEEauthorrefmark{2},
        Ali Nawaz\IEEEauthorrefmark{2}}
    \IEEEauthorblockA{
        \IEEEauthorrefmark{1}AI Team, Doaz Corporation, Seoul 06247\\
        youngtae.kim, soonju.jeong, sardar.arslan, agnidhananjay, yahya.ahmed, ali.nawaz @doaz.ai
    }\\
    \IEEEauthorblockN{
        Jinhee Song, 
        Hyewon Kim}
    \IEEEauthorblockA{
        \IEEEauthorrefmark{2}R\&D Institute, Lotte E\&C, Seoul 06527 \\
        jinhee.song, kim.hyewon @lotte.net
    }
}

\maketitle
\begin{abstract}
This study proposes a two-phase methodology for detecting and classifying auxiliary insulation in structural components. In the detection phase, a YOLOv8x model is trained on a dataset of complete structural blueprints, each annotated with bounding boxes indicating areas that should contain insulation. In the classification phase, these detected insulation patches are cropped and organized into two classes—present or missing—to train a YOLOv8x-CLS model that determines the presence or absence of auxiliary insulation. Preprocessing steps for both datasets involved annotation, augmentation, and appropriate cropping of the insulation regions. The detection model achieved a mean Average Precision (mAP) score of 82\%
%with an accuracy of 91\%
, while the classification model attained an accuracy of 98\%. These findings demonstrate the effectiveness of the proposed approach in automating insulation detection and classification, thereby laying a strong foundation for further enhancements in this domain.
\end{abstract}


\section{Introduction}
\hspace*{1em}Auxiliary insulation plays a critical role in ensuring both the efficiency and safety of industrial systems by minimizing heat loss, preventing condensation, and maintaining overall structural integrity. Despite its importance, verifying the presence and completeness of insulation remains a significant challenge due to the labor-intensive nature of manual inspections. These processes are not only time-consuming but also prone to human error and inconsistencies, leading to overlooked defects that can result in severe operational inefficiencies or safety hazards. As industries move toward the integration of automation and artificial intelligence (AI) in quality assurance, the demand for scalable, precise, and automated insulation defect detection methodologies has grown substantially.

Traditional computer vision techniques have long been explored for defect detection, yet they frequently struggle with the high variability in structural designs, inconsistencies in image acquisition conditions, and the intricate nature of insulation patterns within complex industrial layouts. Variability in blueprint formatting, occlusions, and differing insulation materials further compound the difficulty of generalizing rule-based or conventional feature-extraction methods. With recent advancements in deep learning, particularly through convolutional neural networks (CNNs) and cutting-edge object detection architectures, automated defect detection has reached new levels of reliability and adaptability. Among such architectures, the YOLO (You Only Look Once) model family has demonstrated exceptional performance across various object detection and classification tasks, making it an ideal candidate for insulation defect analysis.

In this study, we introduce an advanced, two-phase deep-learning pipeline designed to automate the detection and classification of auxiliary insulation in industrial environments. In the initial detection phase, we utilize YOLOv8x, a state-of-the-art object detection model, to accurately locate insulation regions within structural blueprints. Following this, the classification phase employs YOLOv8x-CLS to determine whether the identified insulation components are fully present, partially missing, or absent. To further improve the reliability of our model, we extensively augmented our dataset, increasing its size by more than threefold, ensuring robustness against variations in blueprint styles, annotation inaccuracies, and environmental inconsistencies. By combining large-scale dataset augmentation with high-precision deep-learning models, our approach significantly enhances accuracy and efficiency, providing a scalable alternative to manual verification.

To validate the effectiveness of the proposed methodology, we conducted an extensive evaluation on a rigorously curated dataset. Our method was tested against a diverse range of industrial blueprints, capturing insulation structures across different domains. Furthermore, we collaborated with a team of over 10 industry experts, who rigorously reviewed, revised, and cross-verified the model outputs to ensure reliability. This collaborative effort mirrors the quality control processes found in large-scale industrial operations, further reinforcing the credibility of our approach.

This paper provides an in-depth exploration of the dataset preparation process, the experimental setup, and the empirical performance of our methodology. By leveraging advanced deep-learning architectures and a significantly expanded dataset, we demonstrate the potential of AI-driven automation in enhancing insulation defect detection, paving the way for improved industrial quality assurance and safety standards.
%\begin{figure}[t]
%\centering
%\includegraphics[width=30pc]{mouse}
%\caption{(a) Diagram of the experimental set up. (b) Detail of the test object used.}
%\label{fig_env1}
%\end{figure}

%\begin{figure}[t]
%\centering
%\includegraphics[width=21pc]{mouse}
%\caption{Lasing was obtained in the 46.9 nm 3s 1P1 ? 3p 1S0  transition of neon-like Ar by exciting Ar filled alumina capillaries 3.2 mm in diameter with a current pulse having an amplitude Lasing was obtained in the 46.9 nm 3s 1P1 ? 3p 1S0  transition of neon-like Ar by exciting Ar filled alumina capillaries 3.2 mm in diameter with a current pulse having an amplitude}
%\label{fig_env2}\vspace*{-6pt}
%\end{figure}

\section{Related Work}

Research on the automated detection of insulation in architectural and industrial settings remains relatively underexplored. However, related areas such as \textit{blueprint analysis, façade inspection, and thermal defect identification} have seen increasing attention, largely driven by advances in \textit{deep learning} and \textit{computer vision}. Many of these investigations rely on \textit{convolutional neural networks (CNNs)} and \textit{object detection models}, including \textit{YOLO variants}, to identify \textit{structural elements, construction symbols, or drafting errors} within CAD diagrams. While these studies contribute valuable insights into automated detection in construction and industrial applications, they fall short of addressing the specific challenges associated with \textit{auxiliary insulation detection} from \textit{high-resolution blueprint imagery}.  

Several prior works have demonstrated the effectiveness of \textit{object detection models} in analyzing architectural blueprints. For instance, \textit{Kim et al.}~\cite{kim2020automated} explored the use of \textit{deep learning-based object detection} techniques for \textit{automated identification of structural components} such as beams and ducts within construction blueprints. Their study showcased the potential of CNN-driven models in \textit{reducing manual effort} and improving accuracy in \textit{blueprint interpretation}. Similarly, \textit{Ahmed et al.}~\cite{ahmed2020automatic} leveraged CNN-based methods to \textit{detect and localize standard construction symbols} in CAD drawings. Their findings highlighted the \textit{importance of high-resolution input data and rigorous annotation strategies}, reinforcing the need for robust dataset curation when training detection models for architectural applications.  

In a more specialized approach, \textit{Juniat et al.}~\cite{juniat2022automated} proposed an \textit{automated drafting error detection system} based on CNNs. Their work focused on identifying inconsistencies in \textit{architectural CAD symbols}, demonstrating how \textit{symbol recognition models} can be adapted to detect \textit{design inconsistencies and drafting errors}. While their model successfully addressed common blueprint irregularities, it did not extend to identifying \textit{material defects} or evaluating \textit{insulation-related components}, which are the focus of our work.  

In parallel to blueprint-based studies, several researchers have explored \textit{thermal imaging-based insulation defect detection}. \textit{Avdelidis et al.}~\cite{avdelidis2019infrared} and \textit{Pintus et al.}~\cite{pintus2021deep} introduced \textit{deep learning-based two-phase pipelines} that analyze \textit{infrared thermography data} to detect \textit{insulation defects and thermal bridges} in building facades. These methods first \textit{isolate regions exhibiting abnormal heat signatures}, followed by \textit{classification of the defect severity}. Despite their effectiveness in thermal anomaly detection, these approaches are \textit{not applicable to blueprint-based insulation verification}, where defects must be inferred from schematic representations rather than real-world thermal emissions. Nonetheless, these studies underscore the \textit{advantage of combining detection and classification phases}, a strategy we also adopt in our work.  

Unlike prior methods that focus on detecting \textit{generic construction elements}, \textit{drafting errors}, or \textit{thermal anomalies}, our approach introduces a \textit{dedicated solution for auxiliary insulation detection from CAD-derived blueprints}. We propose a \textit{two-phase deep-learning pipeline}, leveraging \textit{YOLOv8x for detection} and \textit{YOLOv8x-CLS for classification}, specifically designed to \textit{localize and evaluate insulation regions} in construction diagrams. Our method surpasses existing approaches in the following key aspects:  

\begin{itemize}
    \item \textbf{Blueprint-Specific Insulation Detection:} Previous studies on blueprint analysis have focused on detecting \textit{structural components or symbols}, whereas our method explicitly \textit{identifies and evaluates insulation placement} within CAD drawings.  
    \item \textbf{Integration of Detection and Classification:} Unlike \textit{symbol detection} or \textit{drafting error analysis}, our pipeline follows a \textit{two-phase workflow} that first \textit{detects insulation regions} and then \textit{classifies them as present or missing}, similar to thermal defect analysis but adapted for blueprint interpretation.  
    \item \textbf{High-Resolution Blueprint Processing:} Our model is designed to process \textit{large-scale construction blueprints} at high resolutions (up to \textit{4800 pixels}), ensuring \textit{detailed pattern recognition} without sacrificing precision.  
    \item \textbf{Dataset Scale and Expert Validation:} A critical limitation in prior research is the \textit{lack of large, verified datasets}. In contrast, our dataset underwent extensive augmentation, expanding from \textit{900 images to over 4,200 detection samples and 23,000 classification patches}. Moreover, \textit{over 10 industry experts} were involved in \textit{revising, reviewing, and double-checking} the labeled dataset, ensuring \textit{annotation accuracy and professional validation}, a step often overlooked in previous studies.  
\end{itemize}  

By combining \textit{state-of-the-art object detection models}, \textit{robust dataset augmentation}, and \textit{expert-driven validation}, our methodology presents a \textit{scalable and industrially viable solution for insulation verification}. In contrast to existing approaches that either \textit{focus on generic structural elements} or \textit{thermal defect detection}, our work directly addresses the challenge of \textit{auxiliary insulation detection in blueprint data}, making it a unique and significant contribution to \textit{computer vision applications in construction and industrial engineering}.

\section{Dataset}
\hspace*{1em} The dataset used in this study was sourced from our industrial partner, \textbf{Lotte E\&C}, and initially consisted of highly detailed Computer-Aided Design (CAD) files representing various construction blueprints. These blueprints contained crucial structural and insulation-related information, but because CAD files are not inherently suited for direct processing through deep-learning pipelines, a multi-stage transformation was necessary before annotation. The raw CAD files were first converted to PDF while applying specific line-thickness settings to optimize readability. Subsequently, these PDFs were exported as high-resolution JPEG images, ensuring consistency in rendering and preparing them for systematic labeling and processing.
\subsection{Data Composition}
The original dataset consisted of 900 high-resolution images, carefully curated to encompass a broad range of insulation conditions. To ensure the model could handle real-world complexities, we categorized these images into three distinct conditions:
 
\begin{itemize}
\item \textbf{Present} : These images represent fully insulated regions, where insulation was correctly installed in every designated area. By incorporating such images, the model learns to establish a baseline for recognizing proper insulation placement, which is critical for reducing false positives in detection.\\
\item \textbf{Missing} : These images depict completely uninsulated areas, where insulation was entirely absent in regions where it was expected. By training on this extreme scenario, the model develops a strong ability to detect cases of total insulation omission, enhancing reliability in real-world deployments.\\
\item \textbf{Partially Missing}: A crucial subset of images contained mixed conditions, where some regions were insulated while others were missing insulation. This reflects real-world inconsistencies, where insulation may have been partially installed or mistakenly omitted in certain areas. Training the model on such variations ensures robustness in detecting partial defects and distinguishing ambiguous cases.
\end{itemize}
By including a balanced mix of all three types, we created a dataset that captures the entire spectrum of insulation defects, ranging from completely insulated to entirely uninsulated, with nuanced partial cases in between. This diversity was essential for constructing a generalizable and adaptable detection model capable of handling varied real-world blueprint conditions.

\begin{figure}[htbp]
\centering
\includegraphics[width=0.5\linewidth]{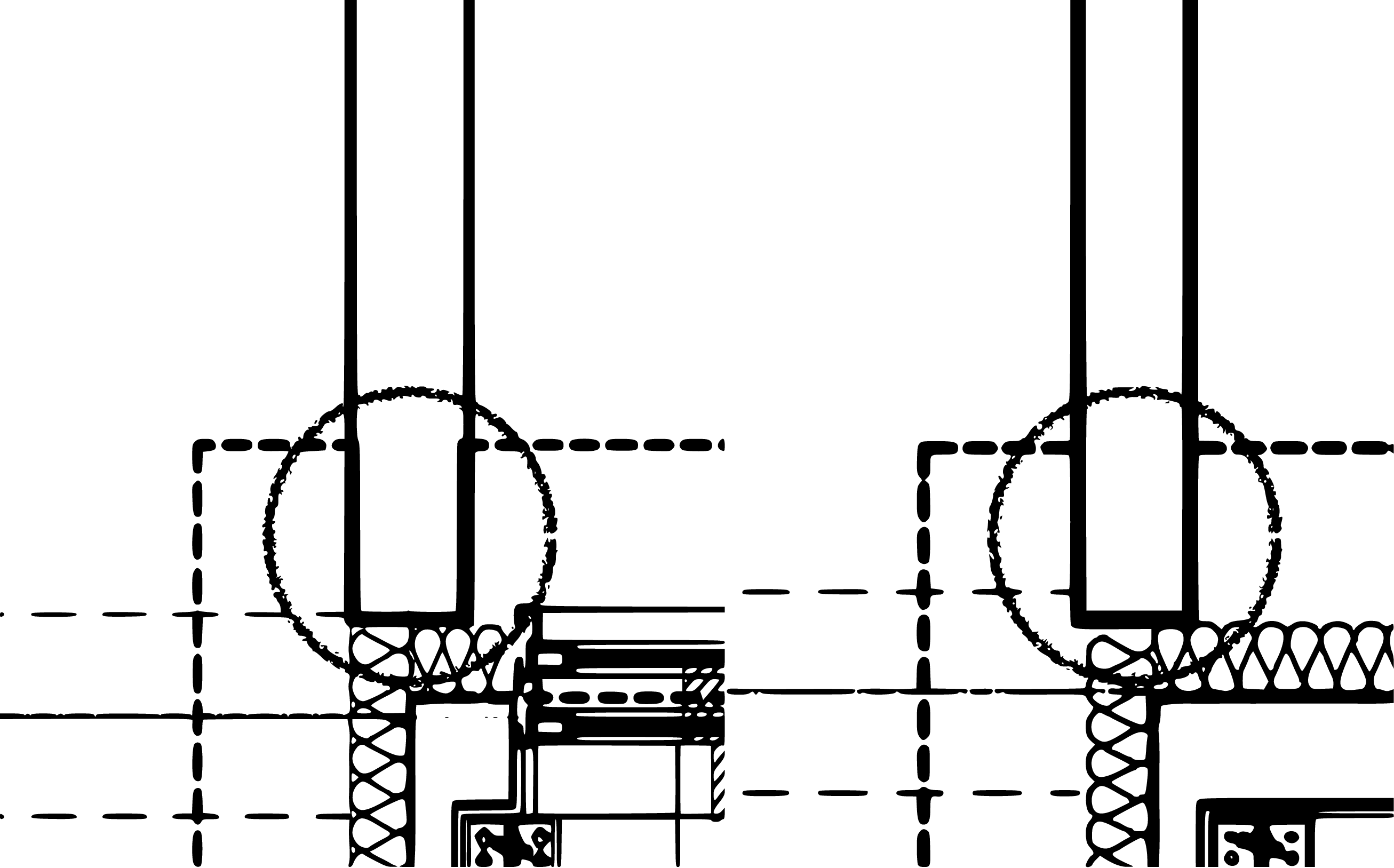}
\caption{(a) Status of present insulation\\\hspace*{1em}\hspace*{1em}\hspace*{1em}\hspace*{1em}\hspace*{1em}(where auxiliary insulation is indicated with thick lines in the marked circular areas.) \\
\hspace*{1em}\hspace*{1em}\hspace*{1em} (b) Status of missing insulation}
\end{figure}

\subsubsection{Data Preparation}
To simulate these three scenarios, we made modifications to DWG files using AutoCAD, carefully adjusting the structural properties to ensure accurate annotations. The DWG files were then exported as PDFs, maintaining controlled line settings to standardize representation across blueprints:
    \begin{itemize}
        \item Structure lines: 0.18-0.25 mm thickness.
        \item Primary insulation: 0.05-0.09 mm thickness.
        \item Auxiliary insulation: 0.09-0.13 mm thickness.
    \end{itemize}
After conversion, the PDFs were further transformed into high-resolution JPEG images to ensure uniformity and clarity for annotation. This multistep process (DWG PDF JPEG) guaranteed consistent line visibility, reducing potential ambiguities during labeling, and allowing for precise identification of insulation-related features.
\subsection{Detection Dataset}
For the detection phase, we labeled a single class—insulation area, which identifies all regions where insulation should exist. Annotations were performed using Roboflow, and the dataset was formatted specifically for YOLOv8, ensuring seamless integration into the object detection pipeline.

To optimize the dataset for model training, we performed focused cropping, removing unnecessary margins and emphasizing the main regions of interest. The cropping specifications were as follows:
\begin{itemize}
    \item \textbf{Cropping :} Width from 5\% (left) to 85\% (right), Height from 10\% (top) to 95\% (bottom).
    \item \textbf{Augmentation :} The dataset was augmented via flipping (vertical) and rotations (clockwise and counterclockwise) to capture variations. Post-augmentation, the detection dataset increased from 900 original images to approximately 3600 images.
\end{itemize}
Following augmentation, the detection dataset expanded from 900 original images to over 4,200 images, significantly increasing the dataset volume and improving the model's ability to generalize across different blueprint conditions.

\begin{figure}[h]
    \centering
    \includegraphics[width=0.5\linewidth]{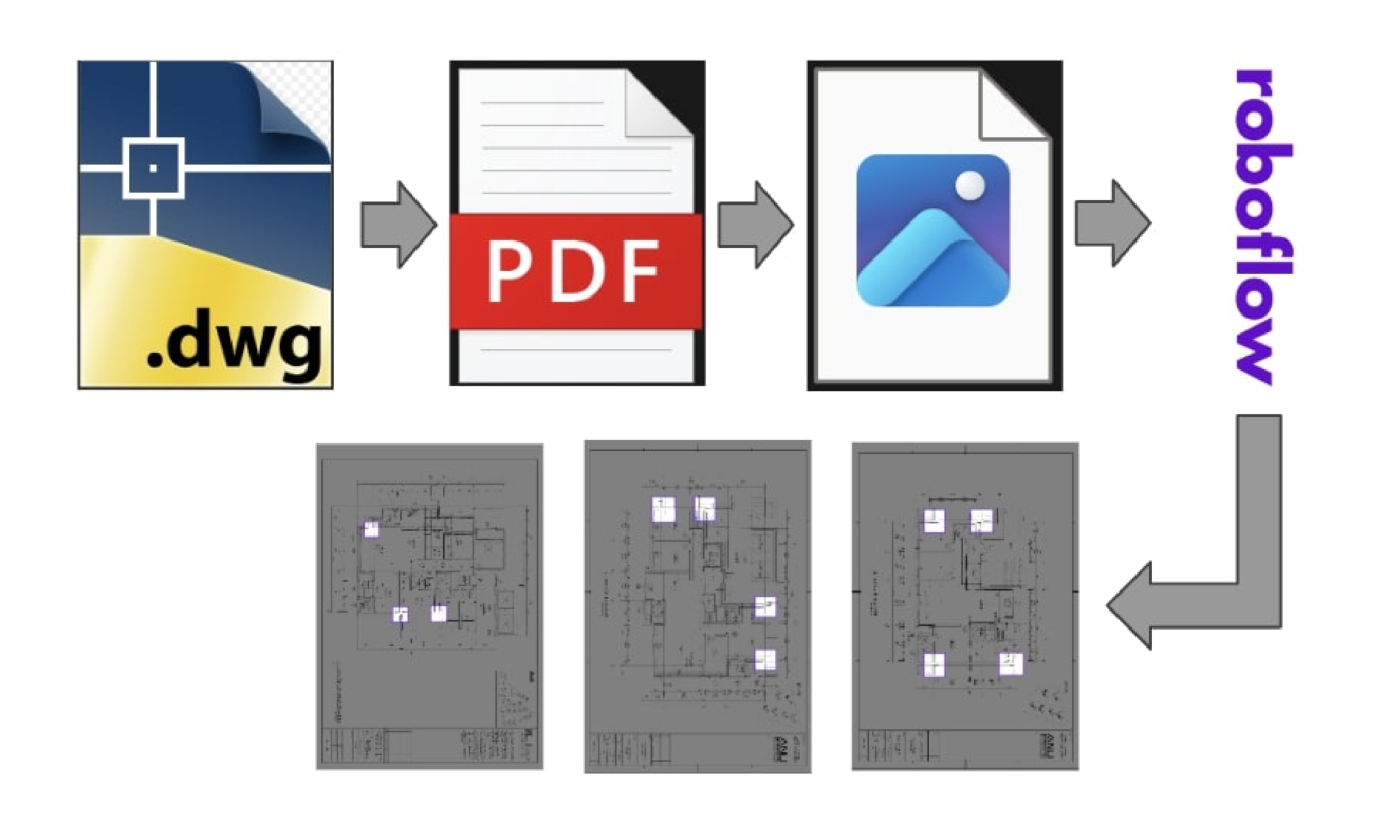}
    \caption{Data preparation workflow for detection model}
    \label{fig:enter-label}
\end{figure}

\subsection{Classification Dataset}
\hspace*{1em} The classification phase was designed to determine whether insulation was present or missing in the detected regions. Using bounding boxes from the detection phase, insulation areas were cropped and categorized into two classes:
\begin{itemize}
    \item Present: Insulation was correctly installed in the cropped region.

        \item Missing: The expected insulation was entirely absent.
\end{itemize}
\begin{figure}[h]
    \centering
    \includegraphics[width=0.5\linewidth]{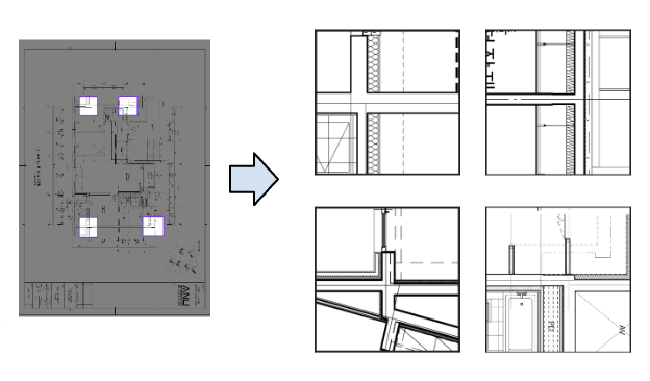}
    \caption{Image of cropped insulation areas from detection data}
    \label{fig:enter-label}
\end{figure}

\hspace*{1em}This approach ensured a structured and systematic classification task, allowing the model to focus on refining its understanding of the presence of insulation within localized blueprint sections.

Through this process, we generated a total of 23,000 cropped patches, significantly larger than the original data set, ensuring that the classification model was trained in a wide variety of cases. These cropped patches were evenly distributed in 11,500 "Missing" samples 11,500 "Present" samples
Augmentation was also applied at this stage, following similar flipping, rotation, and contrast adjustment techniques as in the detection dataset. This not only increased the volume of the dataset but also enhanced the model's robustness against variations in the design styles of the blueprint and environmental noise.

By structuring the dataset into these two distinct phases, we effectively separated insulation localization (detection) from insulation status assessment (classification). This structured methodology, coupled with careful preprocessing and dataset expansion, ensured that both models received high-quality, well-balanced training data, leading to more precise and reliable insulation defect detection in industrial applications.

\subsection{Expert Verification and Quality Assurance}
To ensure the highest level of accuracy and reliability in our dataset, a rigorous post-labeling verification process was conducted. After the initial dataset labeling was completed, a team of over 10 drawing experts was assigned to review, revise, support, and double-check each annotated blueprint. These professionals, with extensive experience in architectural and industrial insulation design, meticulously examined the labeled insulation regions to validate their correctness, ensuring that both detection and classification phases were grounded in highly accurate data.

The scale of this verification process was equivalent to the workforce investment of an entire company, reflecting our commitment to delivering a robust and professional solution. Each expert contributed to multiple rounds of cross-validation, systematically refining the annotations to eliminate potential inconsistencies and ambiguities. This collaborative effort mirrored the quality assurance protocols found in large-scale industrial projects, reinforcing the integrity of our methodology and the dependability of the dataset.

By implementing this multi-expert verification approach, we ensured that our model was trained on a dataset of exceptional quality, minimizing potential biases and errors. This rigorous validation process not only improved the accuracy of insulation detection, but also solidified the credibility of our approach, positioning it as a highly reliable solution for industrial applications. The following section presents the empirical results obtained using this meticulously verified dataset, demonstrating its effectiveness in real-world insulation defect detection scenarios.

\section{Methodology}
\hspace*{1em}In this study, we propose a two-phase deep-learning pipeline designed to automate the detection and classification of auxiliary insulation within structural blueprints. This methodology is structured to first identify insulation regions within large-scale blueprint images using an advanced object detection model. Following this, a secondary classification model determines whether each detected region contains the required insulation. This two-stage approach ensures both accuracy and computational efficiency by leveraging state-of-the-art deep-learning architectures. By separating localization from classification, we enable a precise yet scalable framework capable of identifying insulation defects with minimal manual intervention.

\begin{figure}[h]
    \centering
    \includegraphics[width=1.0\linewidth]{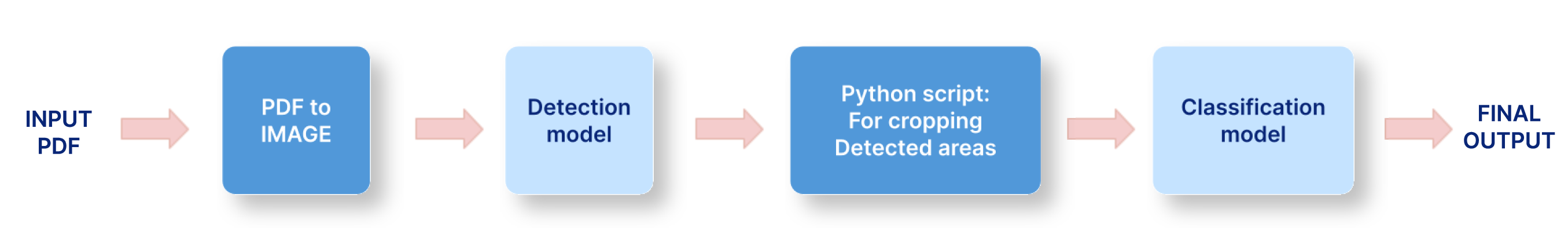}
    \caption{The workflow diagram of the models}
    \label{fig:enter-label}
\end{figure}
\subsection{Detection Phase}
\subsubsection{Model and Training Procedure
}
\hspace*{1em}The first phase of our methodology focuses on detecting insulation regions within blueprints using YOLOv8x, a highly optimized object detection model known for its robust performance on high-resolution imagery. Unlike conventional detection models, YOLOv8x is designed to efficiently process large-scale images while preserving structural detail. Instead of initializing the model with publicly available pretrained weights, we implemented a custom YAML configuration tailored specifically to the insulation detection task. Preliminary experiments revealed that generic pretrained models struggled to converge effectively in this domain, necessitating a domain-specific training approach.

The training process was conducted on a high-performance system equipped with eight NVIDIA A100 GPUs, each with 80 GB of memory, allowing for distributed training across multiple devices. A batch size of 16 was allocated across the GPUs, with each processing two images per iteration. Given the intricate nature of blueprint designs, we set the input resolution to 4800 pixels along the longer edge to capture global structural patterns relevant to insulation placement. The training process employed Stochastic Gradient Descent (SGD) with an initial learning rate of 0.01, as this configuration demonstrated superior convergence compared to alternatives such as Adam. To mitigate overfitting, ann inherent risk given the similarity of images within the data set, a, a dropout rate of 0.2 was applied. Additionally, an early stopping mechanism with a patience threshold of 25 epochs was implemented, halting training if validation performance plateaued. The final model was trained over 100 epochs, requiring approximately 16 hours of compute time.

Following training, each input image was processed through YOLOv8x, yielding bounding box predictions corresponding to insulation regions. To refine these predictions, standard non-maximum suppression (NMS) was applied to remove redundant or overlapping bounding boxes, ensuring that the most relevant insulation areas were preserved. These bounding boxes were subsequently used to extract individual image patches, which were prepared for classification in the second phase of the pipeline.

\subsection{Classification Phase}
\subsubsection{Model and Training Procedure
}
\hspace*{1em}Once insulation regions were localized in the detection phase, the second stage of the methodology focused on classification. The objective of this phase was to determine whether insulation was fully present, partially missing, or completely absent in each detected region. To achieve this, we employed YOLOv8x-CLS, a specialized image classification model optimized for high-accuracy binary decision-making. Unlike the detection phase, which involved full-scale blueprint processing, the classification model operated on cropped patches, significantly reducing computational requirements while maintaining high precision.

Model training was performed using four NVIDIA A100 GPUs, each processing a portion of the dataset in parallel. A batch size of 2048 was distributed across the GPUs, ensuring efficient training convergence. Given the smaller size of input patches, they were resized to 320 pixels while preserving essential structural features. As in the detection phase, the model was trained using SGD with an initial learning rate of 0.01, and a dropout rate of 0.2 was maintained to prevent overfitting. Given the more constrained nature of the classification task compared to detection, the model was trained for 300 epochs with a patience threshold of 50 epochs; however, it successfully converged in under one hour.

Once fully trained, the YOLOv8x-CLS model classified each cropped insulation region as either "present" or "missing," providing a clear assessment of insulation status. By structuring the pipeline in this manner, we ensured that computationally intensive detection was performed only once per image, while classification operated on much smaller image patches, optimizing overall efficiency.

\subsubsection{Cropping and Data Flow}

To facilitate a seamless transition between detection and classification, we developed a custom Python script that automated the merging of bounding box predictions with the original blueprint images. This script systematically extracted each predicted insulation region and stored it as an independent patch for classification. By standardizing the cropping process, we ensured consistency across all dataset samples, eliminating manual errors and potential inconsistencies. This automated workflow also improved reproducibility, ensuring that each insulation patch was processed in a controlled and systematic manner.

By structuring the task into two distinct phases, our methodology effectively capitalizes on the strengths of deep-learning models for both localization and classification. The detection phase leverages the power of modern object detection architectures to identify insulation regions, while the classification phase ensures precise determination of insulation presence within those regions. This structured approach enables accurate and scalable insulation defect detection in industrial applications, reducing the reliance on labor-intensive manual inspections while enhancing overall quality assurance processes. The next section presents the results obtained using this methodology, along with a detailed evaluation of its effectiveness in real-world settings.

%\begin{table}
%\caption{This is an example of Table legend.}
%\centering
%\includegraphics{bozku.t1.eps}
%\label{tab1}
%\end{table}

\section{Results}
\begin{table}[h]
\centering
\caption{Performance Metrics for Detection and Classification Models}
\begin{IEEEeqnarraybox}[\IEEEeqnarraystrutmode\IEEEeqnarraystrutsizeadd{2pt}{1pt}]{c/r/c/c}
\IEEEeqnarrayrulerow\\
\textbf{Model} & \textbf{Metric} & \textbf{Value}\\
\IEEEeqnarraydblrulerow\\
\IEEEeqnarrayseprow[2pt]\\
Detection (YOLOv8x) & mAP & 82\%\\
\IEEEeqnarrayseprow[2pt]\\
Classification (YOLOv8x-CLS) & Accuracy & 98\%\\
\IEEEeqnarrayseprow[2pt]\\
\IEEEeqnarrayrulerow
\end{IEEEeqnarraybox}
\end{table}

%%% Detection Results visuals

% \begin{figure}[h]
%     \centering
%     \includegraphics[width=0.5\linewidth, bb=0 0 600 400]{results/image-1.jpg}
%     \caption{Detection model results}
%     \label{fig:enter-label}
% \end{figure}

%%% Both models Results visuals
\begin{figure}[H]
\centering
% First subfigure
\begin{subfigure}[b]{0.45\textwidth}
\centering
\includegraphics[width=\textwidth]{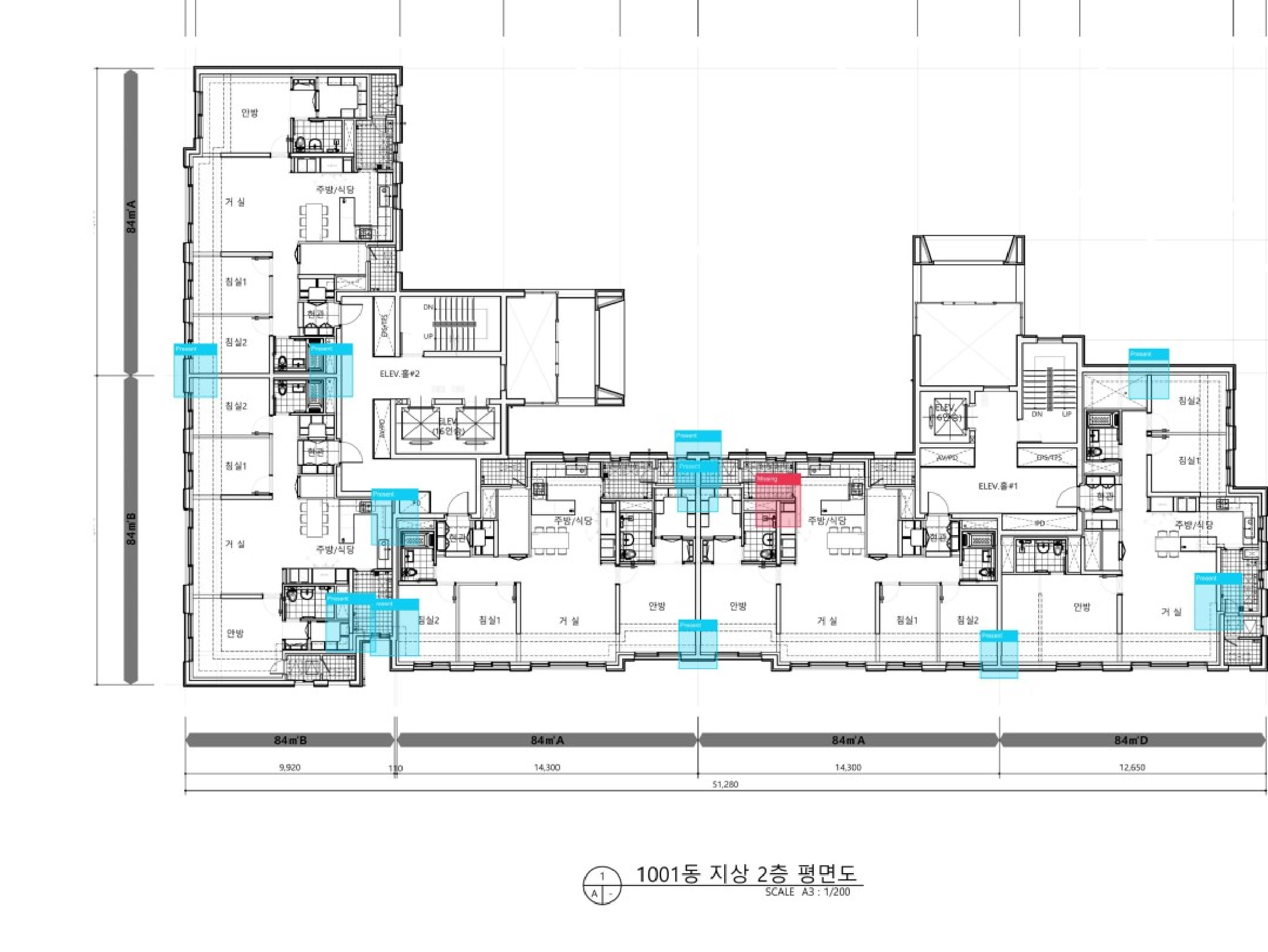} % Replace 'image1' with your filename
\caption{Example 1}
\label{fig:sub1}
\end{subfigure}
\hfill
% Second subfigure
\begin{subfigure}[b]{0.45\textwidth}
\centering
\includegraphics[width=\textwidth]{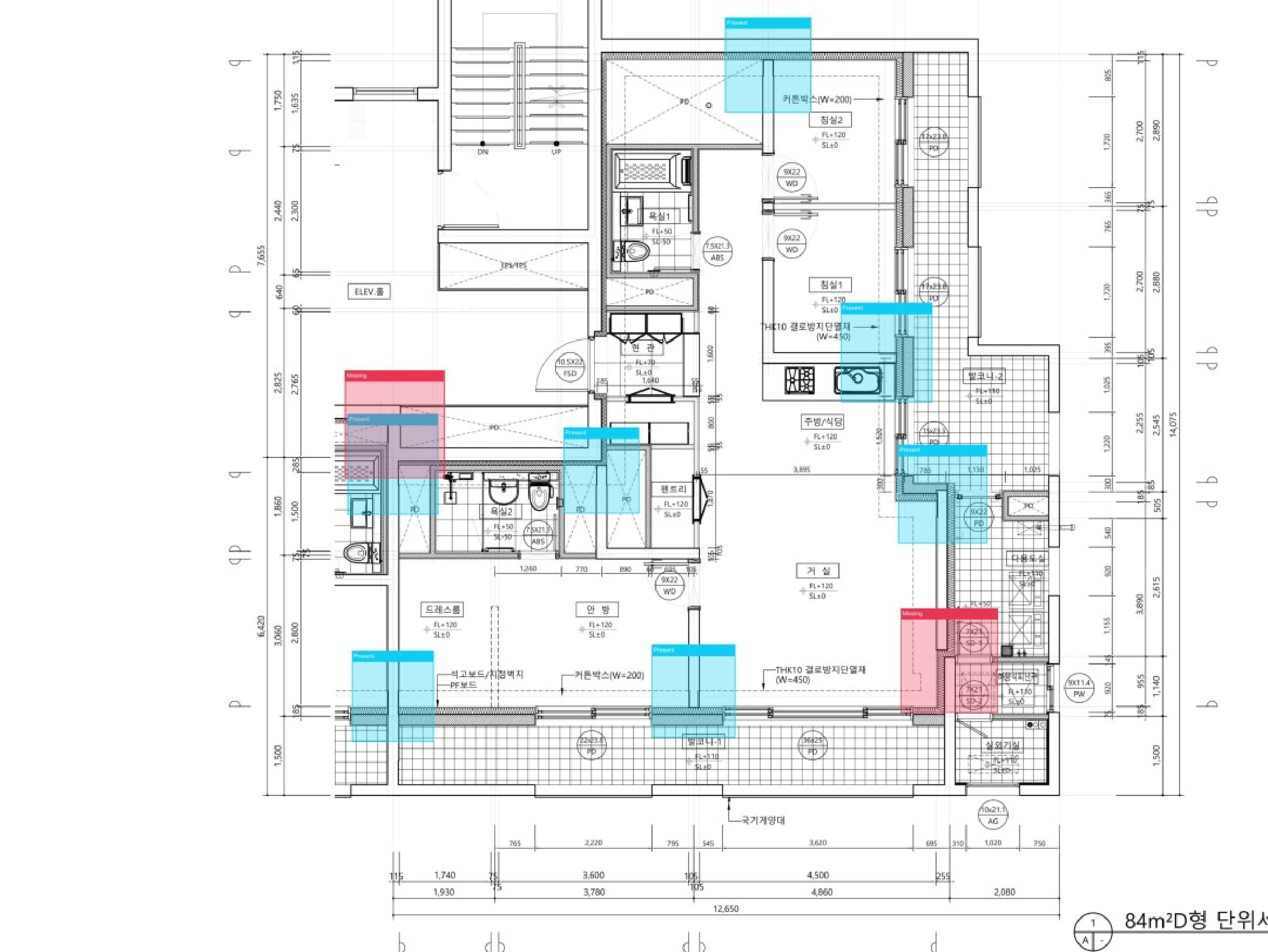} % Replace 'image2' with your filename
\caption{Example 2 }
\label{fig:sub2}
\end{subfigure}

\caption{Model predictions}
\label{fig:two_images}
\end{figure}

\subsection{Detection Model
}
The YOLOv8x detection model was evaluated using a separate validation set held out during training. The evaluation centered on two primary metrics: mean Average Precision (mAP) and accuracy. The model achieved a mean mean precision of 82\%, indicating that it accurately located insulation areas within the blueprints under various testing conditions. Furthermore, it achieved an accuracy of 91\%, reflecting the proportion of correct predictions (true positives and true negatives) relative to all predictions made during the detection phase.
\begin{figure}[htbp]
\centering
% First subfigure
\begin{subfigure}[b]{0.45\textwidth}
\centering
\includegraphics[width=\textwidth]{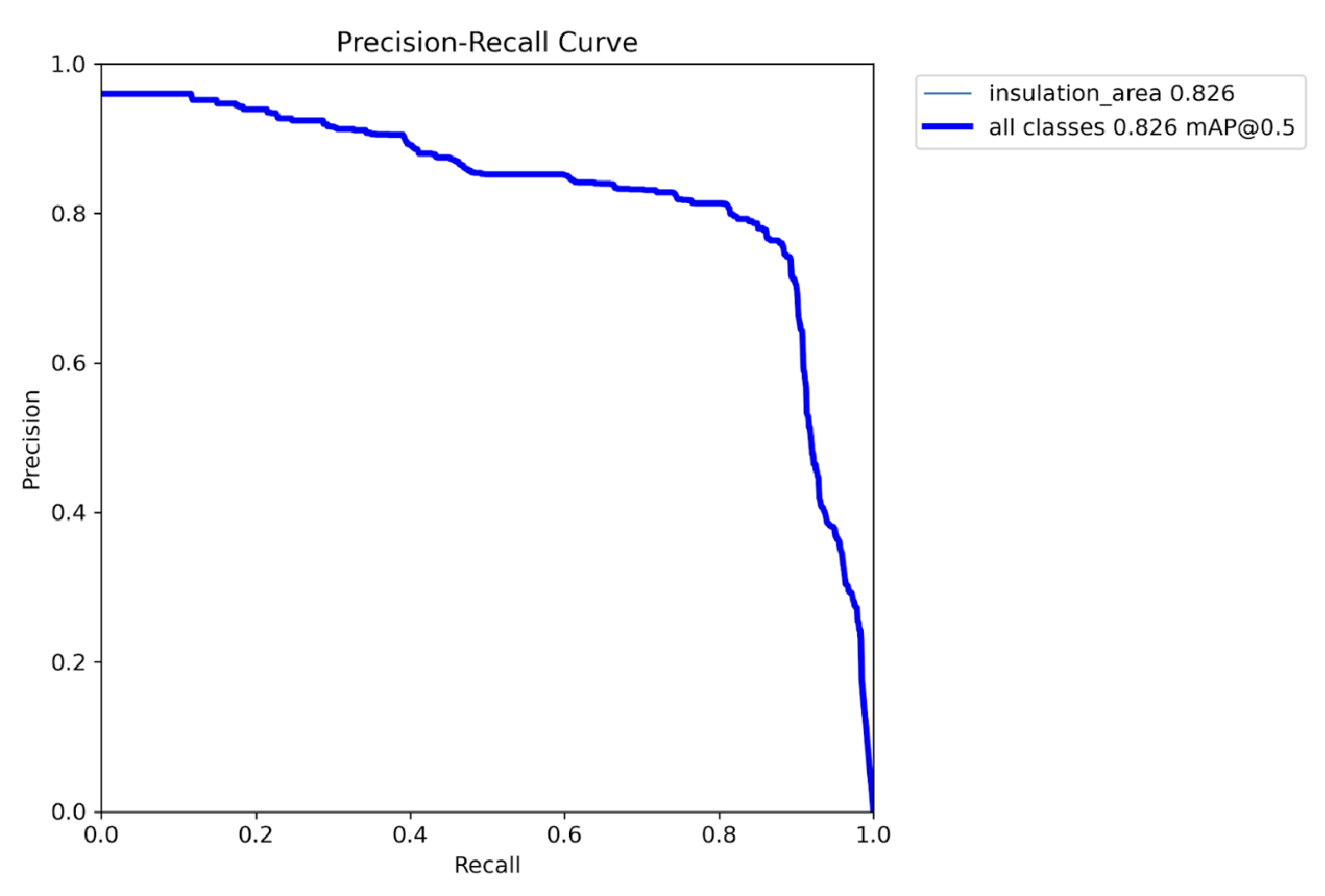} % Replace 'image1' with your filename
\caption{Precision-Recall curve}
\label{fig:sub1}
\end{subfigure}
\hfill
% Second subfigure
\begin{subfigure}[b]{0.45\textwidth}
\centering
\includegraphics[width=\textwidth]{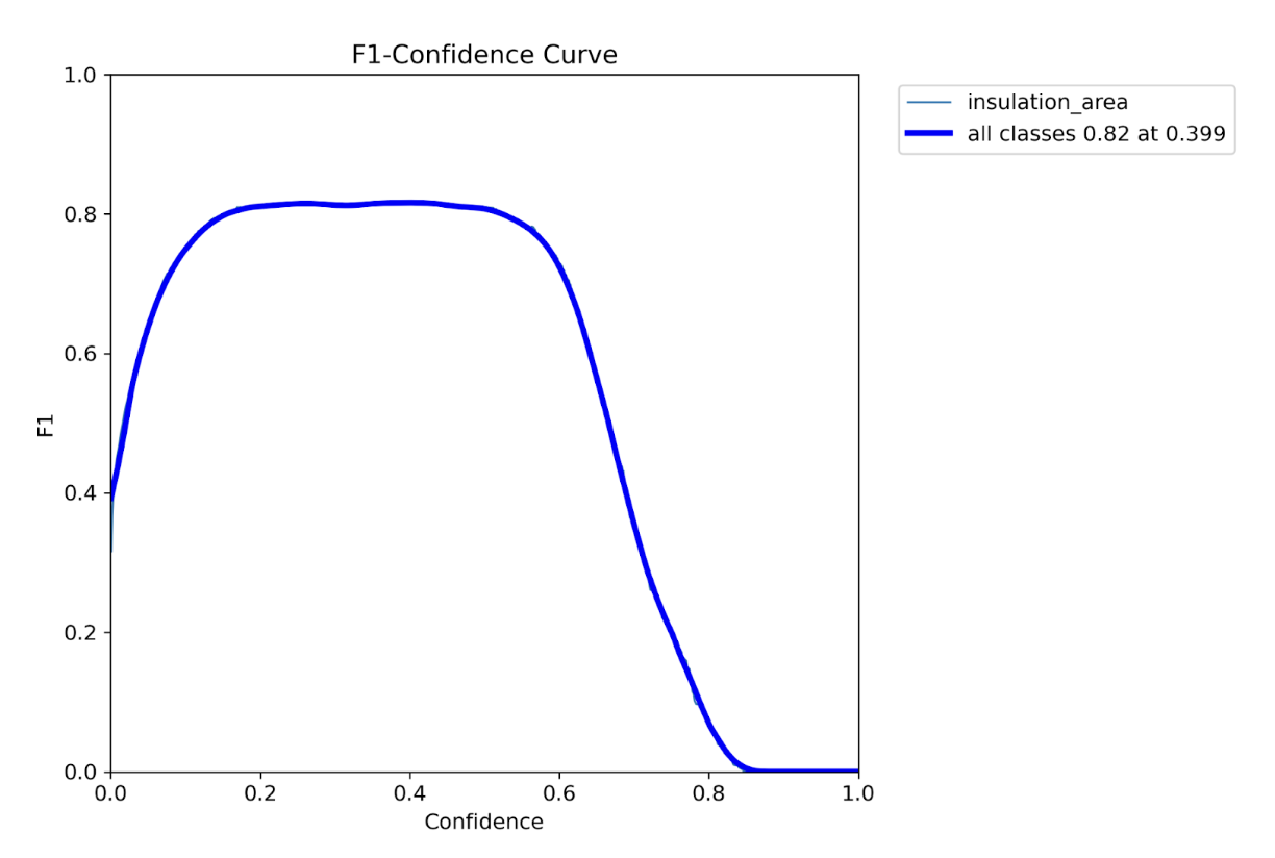} % Replace 'image2' with your filename
\caption{F1 curve }
\label{fig:sub2}
\end{subfigure}

\caption{Detection model metric}
\label{fig:two_images}
\end{figure}
\begin{figure}[h]
    \centering
    \includegraphics[width=1\linewidth]{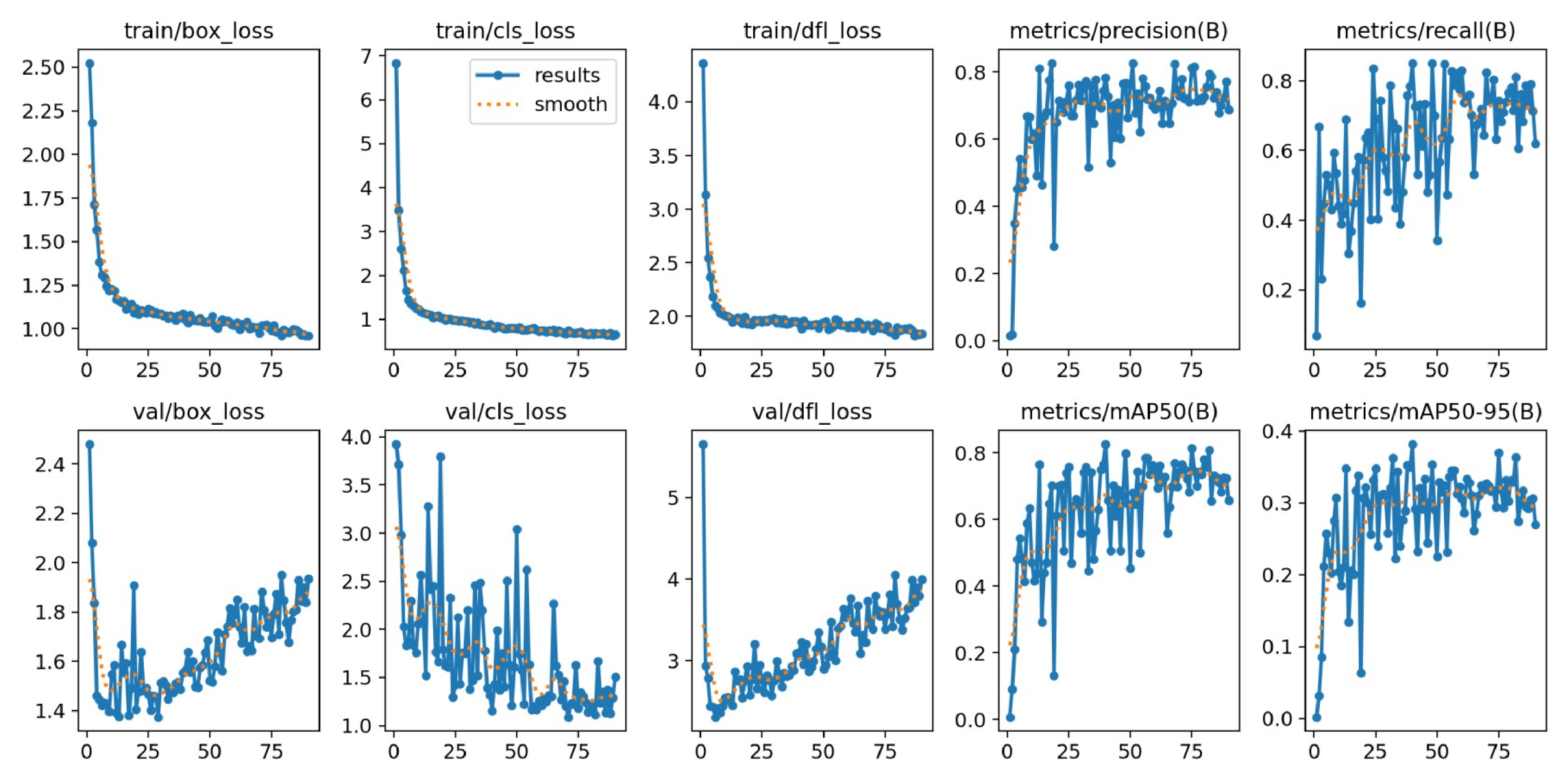}
    \caption{ Other Detection model metrics}
    \label{fig:enter-label}
\end{figure}
\subsection{Classification Model
}
In the second phase, the YOLOv8x-CLS classification model was tested on cropped insulation patches derived from the bounding boxes identified in the detection phase. The primary evaluation metric for this binary classification task (i.e., “present” versus “missing”) was accuracy. The classification model yielded an accuracy of 98\%, suggesting a high degree of precision in differentiating between correctly installed insulation and missing regions.

Overall, these results demonstrate the effectiveness of the proposed two-phase methodology. The strong detection performance ensures that relevant insulation areas are reliably identified within large, complex blueprint images, while the high classification accuracy enables confident determination of whether insulation is present or missing in each localized region. The combined pipeline thus offers a robust framework for automating insulation defect detection in industrial construction settings.

\begin{figure}[h]
    \centering
    \includegraphics[width=0.5\linewidth]{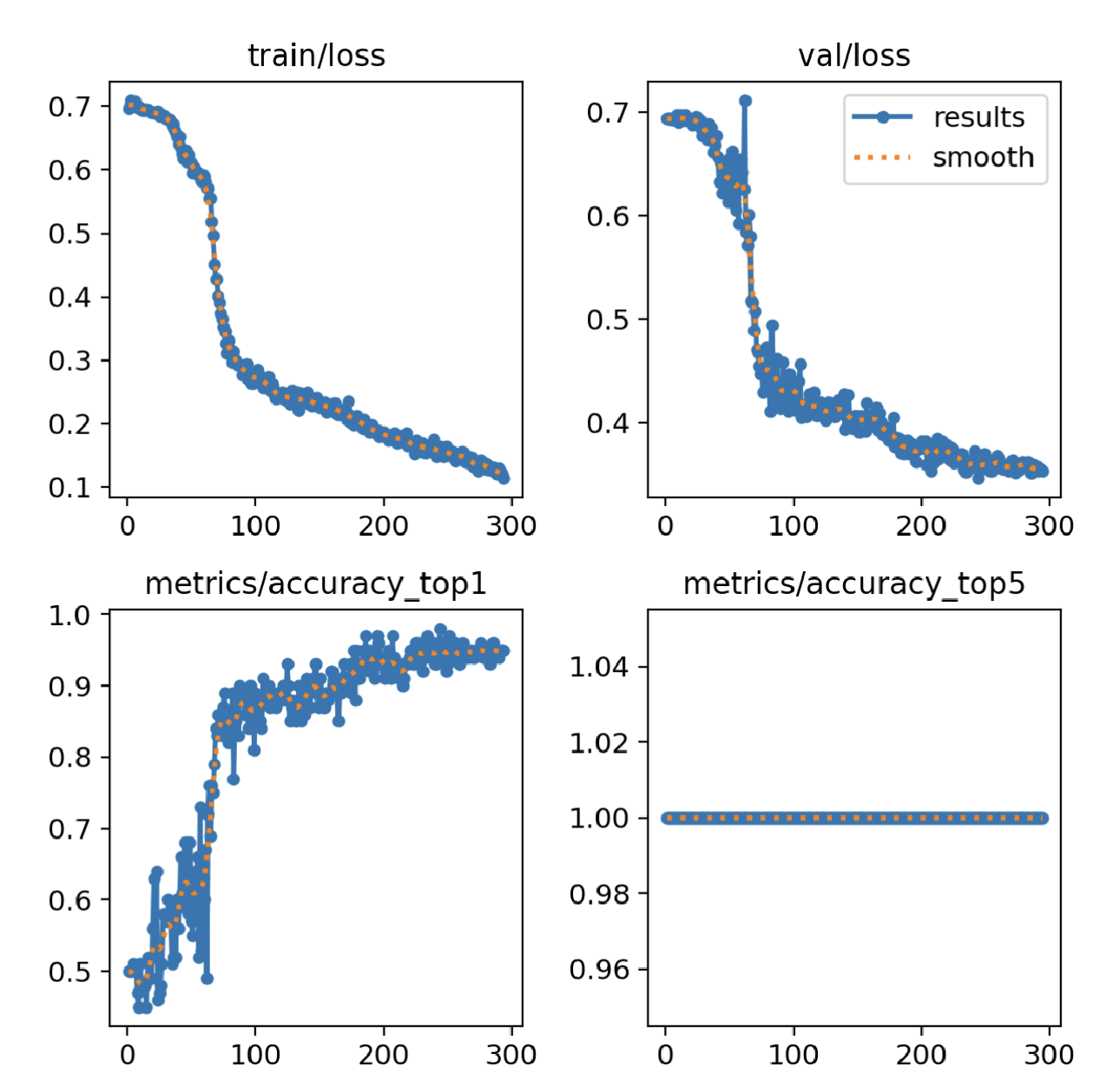}
    \caption{Classification model metrics}
    \label{fig:enter-label}
\end{figure}
%CONFUSION METRICES
\begin{figure}[H]
\centering
% First subfigure
\begin{subfigure}[b]{0.45\textwidth}
\centering
\includegraphics[width=\textwidth]{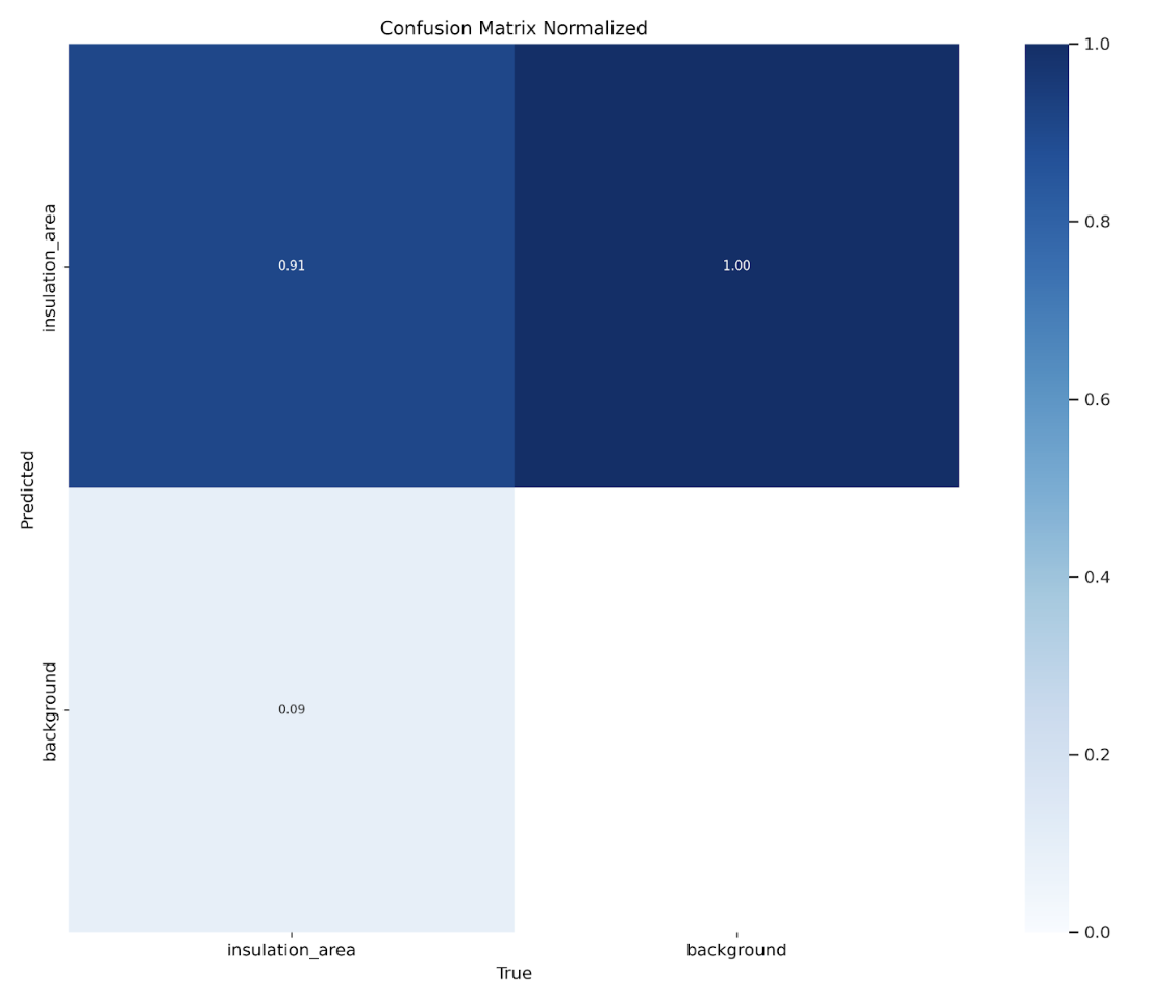} % Replace 'image1' with your filename
\caption{For Detection model}
\label{fig:sub1}
\end{subfigure}
\hfill
% Second subfigure
\begin{subfigure}[b]{0.45\textwidth}
\centering
\includegraphics[width=\textwidth]{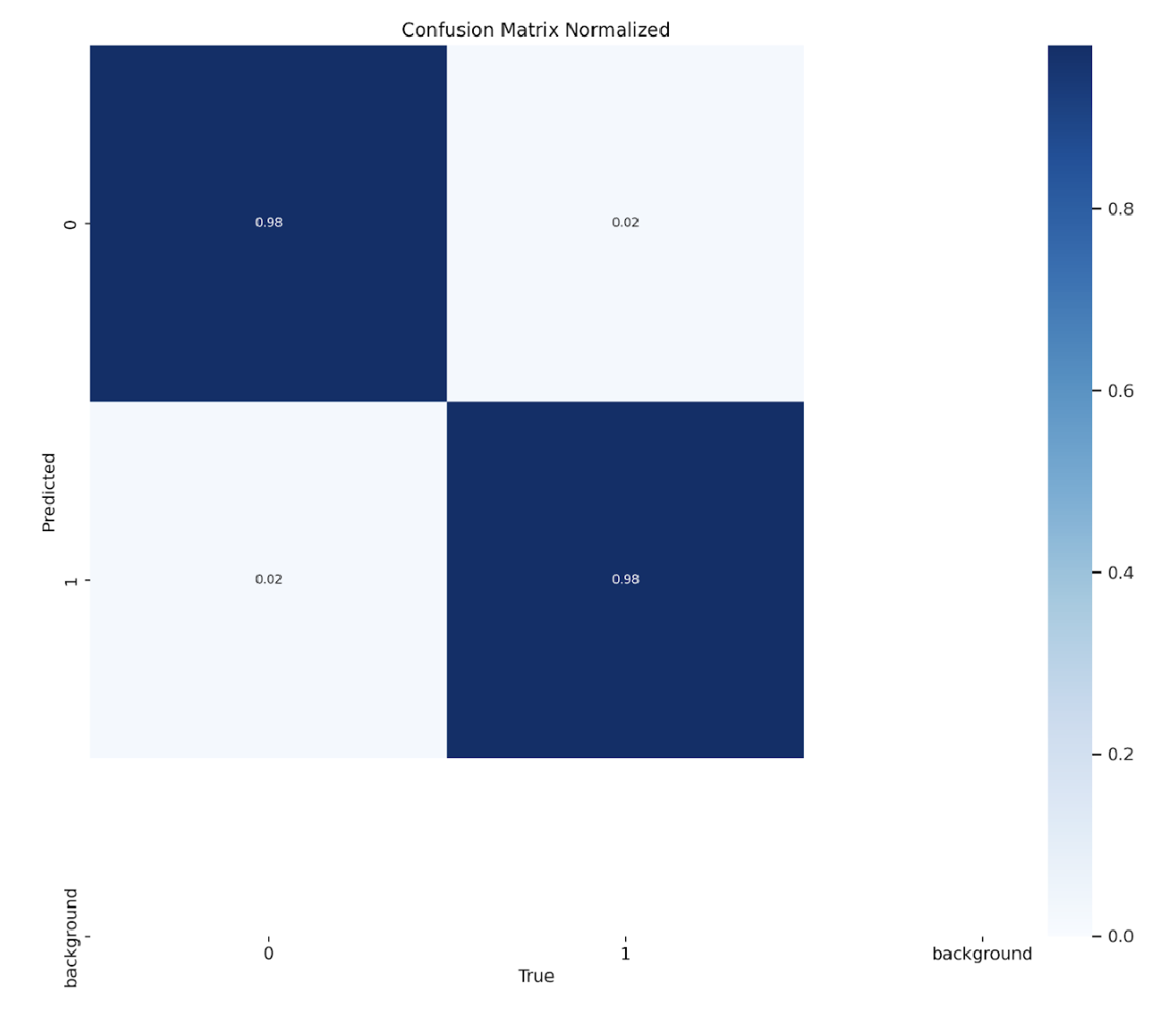} % Replace 'image2' with your filename
\caption{For classification model }
\label{fig:sub2}
\end{subfigure}

\caption{Confusion matrix for both models}
\label{fig:two_images}
\end{figure}
\section{Future Directions}
\label{sec:future_directions}

While the proposed two-phase YOLO-based pipeline has demonstrated significant improvements in the automated detection and classification of auxiliary insulation defects in structural blueprints, several avenues remain for future exploration. Expanding the scope of this research will enable more robust and generalizable insulation verification systems applicable across diverse construction and industrial environments.

One promising direction involves extending the detection methodology to include \textit{cross-sectional views} of structural layouts, in addition to the current analysis of floor plans. Cross-sectional blueprints often contain crucial details regarding insulation placement within walls, ceilings, and mechanical systems that are not fully visible in standard floor plan representations. Integrating a multi-perspective approach will provide a more comprehensive understanding of insulation coverage and potential defects.

Further enhancements can be made by incorporating \textit{multi-modal data fusion}, leveraging additional sources such as \textit{infrared thermography}, \textit{3D point cloud data}, and \textit{Building Information Modeling (BIM)}. By integrating thermal imaging data, for example, the model could cross-validate detected insulation regions with real-world heat distribution patterns, improving defect detection accuracy. Similarly, fusing blueprint data with BIM models would allow for more context-aware insulation verification, ensuring alignment between as-designed and as-built conditions.

Another critical area of development is the expansion of the dataset to improve model generalizability. Although the current dataset was significantly augmented and verified by a team of industry experts, collecting real-world annotated blueprints from multiple construction sites and industrial projects would further enhance robustness. Additionally, developing a \textit{semi-supervised learning framework} that incorporates self-training or active learning techniques could reduce reliance on manually labeled datasets, making large-scale deployment more feasible.

Improving the interpretability of the model outputs is another key area for future work. Current deep-learning approaches, particularly object detection models, often function as black-box systems, making it difficult to explain misclassifications or inconsistencies. Implementing \textit{explainable AI (XAI)} techniques, such as attention visualization and feature attribution methods, would provide greater transparency into model decision-making. This would be particularly valuable in industrial applications where validation from human experts remains essential.

Lastly, there is an opportunity to apply similar methodologies beyond auxiliary insulation detection. The proposed framework can be adapted for detecting \textit{other construction materials}, such as fireproofing layers, acoustic insulation, or even structural reinforcements. Extending the system’s capabilities to additional material verification tasks would broaden its applicability in the construction and industrial automation domains.

By addressing these future directions, the proposed methodology can evolve into a more comprehensive, scalable, and widely applicable solution for automated insulation verification. Continued advancements in deep learning, dataset curation, and multi-modal data integration will pave the way for more intelligent and reliable construction quality assurance systems.

\end{document}